\documentclass[11pt]{article}
\usepackage[letterpaper,margin=1in]{geometry}
\usepackage{amsmath,amssymb}
\usepackage{times}
\usepackage{amsmath}
\usepackage{algorithm}
\usepackage{algpseudocode}
\usepackage{booktabs}
\usepackage{graphicx}
\usepackage{hyperref}
\usepackage{url}
\title{LMLineage: Content Lineage and Context Governance for LLM Agents}
\author{Wei Hou$^{1,*}$, Jingxin Geng$^{1}$, Xin Chen$^{1}$, Zheqing Jiang$^{1}$\\
$^{1}$Lenovo Global Security Lab\\
\texttt{\{houwei6, gengjx2, chenxin34, jiangqz2\}@lenovo.com}\\
Wei Hou also at: \texttt{dr.houwei@gmail.com}\\
$^{*}$Corresponding author}
\date{}

\begin{document}
\maketitle

\begin{abstract}
Interpretability for AI agents should not be confined to understanding the model's internal weights --- it must also cover the agent's reasoning and decision-making process over multi-source context. We present \textbf{LMLineage}, an out-of-band system that attributes every output sentence of an agent session to its supporting input spans with calibrated confidences, without requiring white-box access to the model or any additional inference passes. On top of the lineage metadata, we build \textbf{concept-drift micro-compaction}: instead of waiting for the context-length limit and summarizing everything, the system detects when previously used input documents have stopped being referenced (``drifted'') and surgically compacts only those, preserving user instructions verbatim. We evaluate on a multi-tier corpus: 71 gold-annotated sentences from 10 Chinese web-QA cases, plus 90 turns of independently collected multi-turn sessions across three generator models (gpt-4o-mini, qwen3.8-27b, deepseek-v4-flash). Results show that (i) window-level embedding similarity is far better aligned with LLM paraphrasing than whole-document or lexical matching; (ii) the greedy-window fast path is both faster and more accurate on average than exhaustive line-matching (a multiple-comparisons bias in the latter inflates web recall at the cost of overall accuracy); (iii) per-turn attribution completes in ${\sim}$1\,s on a single GPU, remaining stable as context accumulates to 40 turns; and (iv) concept-drift detection is conservative (0\% false compaction) with recall bounded by attribution accuracy. We further describe the concept-drift micro-compaction design and an interactive incident-tracing/memory-cleaning case study. A companion patent application covers the lineage construction method.
\end{abstract}

\section{Introduction}

LLM agents in production settings consume heterogeneous inputs --- system prompts, user turns, retrieved documents, tool results, and persistent memory --- and produce actions and text whose provenance is opaque. When an agent misbehaves, operators today cannot answer a basic question: \emph{which input sentences caused which output sentences?} This is not merely a debugging inconvenience --- it is a fundamental gap in agent interpretability. Current interpretability research focuses overwhelmingly on the model's internal representations (attention maps, feature attribution, mechanistic analysis), yet agents make decisions through a \emph{process} that spans multiple turns, tool calls, and context transformations. Understanding this process requires knowing, at each step, what information the agent used and where it came from.

Three industry pain points motivate fine-grained \emph{content lineage} for agent sessions: (i) context compaction can silently drop critical constraints --- in one publicly reported incident, an AI assistant with mailbox access deleted business emails after aggressive compaction removed the ``do not delete'' instruction, and there was no objective way to tell what was lost; (ii) memory systems accumulate unverifiable content with no traceable origin, diluting useful memory and degrading later reasoning; (iii) post-hoc incident analysis lacks a sentence-level link between outputs and inputs, making root-cause analysis a manual slog through raw transcripts. All three reduce to the same missing capability: knowing, for each output sentence, which input spans support it and with what confidence.

We argue that an ideal interpretability tool for agents should be: (a) \emph{out-of-band} --- it must not require modifying the agent's design or workflow, nor add inference passes to the critical path; (b) \emph{not GPU-dependent} --- it should run on commodity CPU infrastructure alongside the agent, not compete for the same accelerators; and (c) \emph{process-aware} --- it must capture the agent's reasoning over time, not just static model outputs. LMLineage is our attempt at this vision.

Existing attribution techniques fall short here. Self-citation prompting asks the model to cite its sources, but citations are themselves model generations and can be unfaithful [Gao et al., 2023; Rashkin et al., 2023]. White-box methods based on model internals require multiple inference passes per output token [Qi et al., 2024]. Learned attribution models require task-specific training data. Lexical methods (BM25, edit distance), long the workhorse of plagiarism and near-duplicate detection [Robertson \& Zaragoza, 2009; Christen \& Naumann, 2013], break down under the heavy paraphrasing LLMs produce, as our measurements confirm. And classical data lineage operates at file/table granularity [Buneman et al., 2001], not sentences.
LMLineage takes a different route: \textbf{purely out-of-band, black-box, sentence-to-span attribution}. A sidecar service observes the agent's session events, segments every input document and output sentence, and computes attribution by locating, for each output sentence, the best-matching \emph{word-level window} inside each input document using embedding similarity, then converting the best window into inherited metadata triplets (source, value, confidence). No model re-invocation is needed: attribution for a turn finishes in seconds on a single GPU, in parallel with the next turn.

Because the lineage tells us \emph{which input documents are still being used}, a natural control application emerges: \textbf{concept-drift detection}. When a document stops contributing to new outputs for several consecutive turns, it is likely stale context --- a candidate for surgical removal. We implement this as a micro-compaction path that replaces only the drifted non-user documents with a 1--2 sentence summary, instead of a full-context summarization pass.

\paragraph{Contributions.}
\begin{enumerate}
\item A working out-of-band lineage system for LLM agent sessions (implemented as an OpenClaw plugin + attribution service), with a word-window + embedding attribution algorithm and a confidence-propagation scheme that reserves mass for the model's internal knowledge.
\item A 71-sentence gold-annotated benchmark for Chinese agent web-QA attribution, evaluated at two granularities (coarse input-vs-model and fine source-type). We show that the greedy-window fast path achieves both higher overall accuracy and lower latency than exhaustive line-matching, which suffers from multiple-comparisons bias that inflates web recall at the cost of global precision. The fast path is sufficient for downstream drift detection, while line-match serves only as an on-demand verification tool.
\item The concept-drift micro-compaction mechanism --- to our knowledge the first use of lineage metadata as a \emph{proactive trigger} for context compaction --- with a triage policy and interactive case studies.
\end{enumerate}

\section{Related Work}

\paragraph{Answer attribution for RAG.} Retrieval-augmented generation [Lewis et al., 2020] makes provenance questions unavoidable, and the AIS framework formalizes what it means for a statement to be attributable to a source [Rashkin et al., 2023]. Self-citation prompting asks LLMs to cite sources inline; ALCE [Gao et al., 2023] benchmarks and improves such citations, but citations remain model-generated claims that can be unfaithful. Qi et al. [2024] attribute answers via model internals (token-level KL divergence with/without context, then per-document ablations), which requires white-box access and tens of extra inference passes per answer. LMLineage differs from all of these: it is fully out-of-band (no model access, no extra inference), training-free, and targets sentence$\to$span granularity with a calibrated confidence split that explicitly accounts for the model's parametric knowledge.

\paragraph{Context compression and management for LLMs.} Long-context models still exhibit ``lost in the middle'' degradation [Liu et al., 2024], motivating context compression: LLMLingua [Jiang et al., 2023] and follow-ups prune tokens by perplexity; MemGPT [Packer et al., 2024] manages virtual memory with archival/recall stores; agent frameworks (e.g., OpenClaw) trigger whole-context summarization near the token limit. None of these track \emph{which} inputs remain relevant. Our micro-compaction is complementary: lineage tells us \emph{what} to compress; any summarizer can be the \emph{how}.

\paragraph{Sentence embeddings and lexical matching.} Sentence-transformer embeddings [Reimers \& Gurevych, 2019] provide the semantic similarity at the core of our window search; BM25 [Robertson \& Zaragoza, 2009] and edit-distance variants [Christen \& Naumann, 2013] remain standard lexical baselines in duplicate/plagiarism detection --- we show both fail on LLM-paraphrased text (\S5.2).

\paragraph{Data lineage.} Classical data lineage and database provenance track file/table-level origins [Buneman et al., 2001]. We are not aware of prior work constructing sentence-level, confidence-weighted lineage across the full multi-source input of LLM agent sessions; a companion patent application [Hou et al.] describes the construction method.

\section{System}

\subsection{Architecture}

\begin{figure}[t]
\centering
\includegraphics[width=\columnwidth]{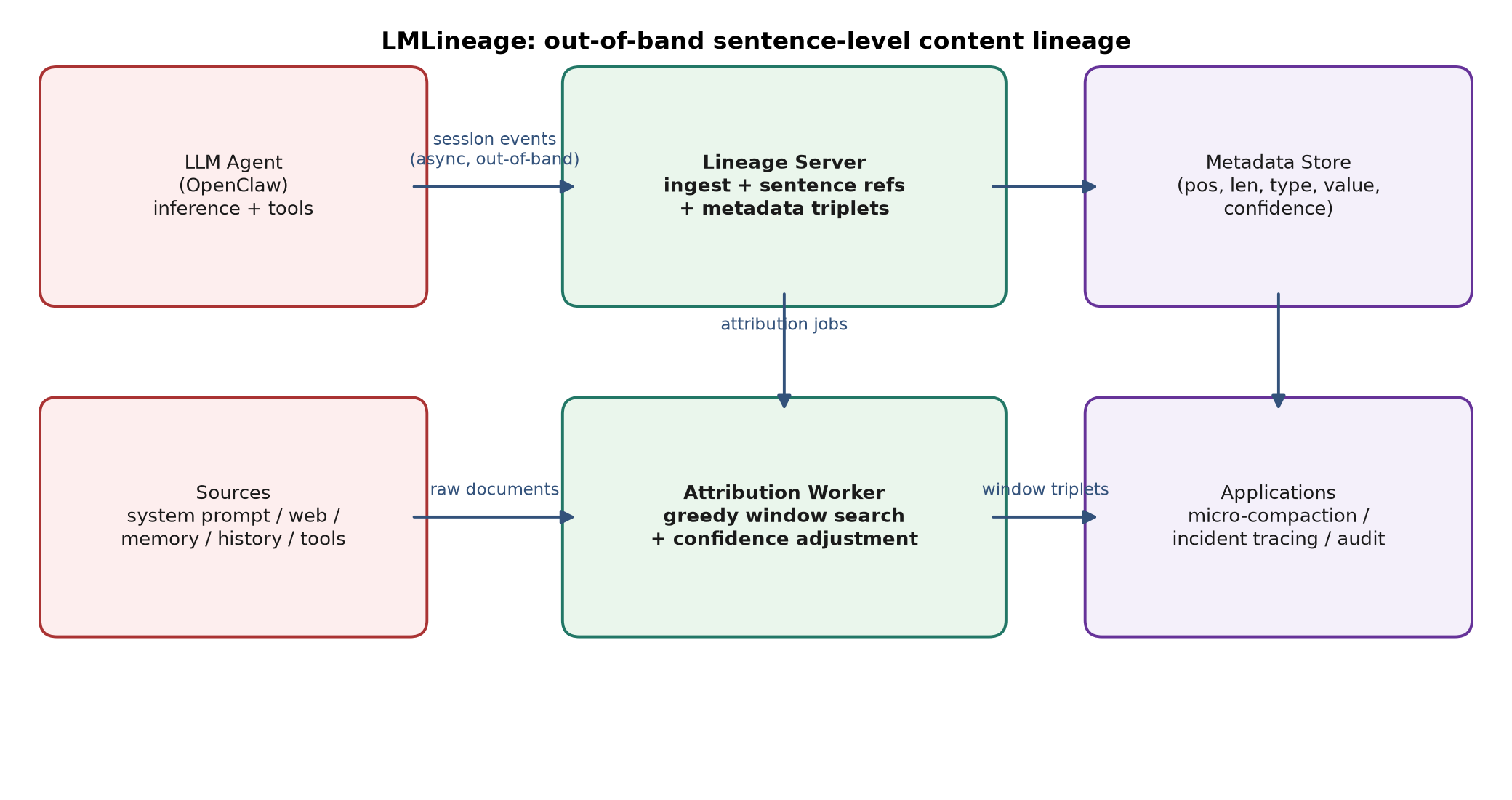}
\caption{System architecture. Session events flow asynchronously from the agent to the lineage server; the attribution worker writes window triplets to the metadata store, which serves downstream applications (micro-compaction, incident tracing, audit).}
\label{fig:arch}
\end{figure}

LMLineage runs \textbf{out-of-band} beside the agent (Fig.~\ref{fig:arch}). An OpenClaw plugin streams session events (turns, retrieved documents, tool results, memory writes) to a lineage server; a worker process attributes each new output sentence to input spans and persists metadata triplets. The agent's inference path is never blocked: attribution is asynchronous, with a readiness API that downstream controllers (e.g., the compaction triage, \S4) can poll with a bounded wait.

\subsection{Attribution algorithm}
\label{sec:algo}

Let $s_i$ denote an output sentence, and let $D = \{d_1, \dots, d_m\}$ denote the multi-source input documents of the turn (web content, system prompt, user prompt, knowledge base, memory, prior context). Each input sentence carries a metadata set of triplets $(\text{type}, \text{value}, c)$ with $c \in [0,1]$ ($c{=}1$ at ingestion; \S\ref{sec:meta}). The objective is to determine, for each $s_i$, which $d_j$ supports it and with what confidence.

Lexical methods (TF-IDF, BM25, edit distance) cannot accommodate cross-lingual content or LLM paraphrase; plain document-level embedding dilutes the supporting span within a long document. We therefore propose a window-first, document-level attribution algorithm (Alg.~\ref{alg:attribution}), consisting of five steps plus a confidence-correction step.

\begin{algorithm}[t]
\caption{Window-First Sentence Attribution}
\label{alg:attribution}
\begin{algorithmic}[1]
\Require Output sentence $s_i$; input documents $D = \{d_1,\dots,d_m\}$; sentence-embedding model $\phi$; word embeddings $\psi$
\Ensure Primary source $d^* \in D$; adjusted triplets $\mathcal{T}_i$
\State \textbf{Word mapping.} Tokenize $s_i$ and each $d_j$; map every token to $\psi$, yielding word--word cosine matrix $M \in \mathbb{R}^{|s_i| \times |d_j|}$
\State \textbf{Greedy window search.} On each $d_j$, find window $w = (pos, len)$ maximizing
\[
\text{score}(w) = \underbrace{\frac{1}{|s_i|} \sum_{t \in s_i} \max_{u \in w} \cos(t, u)}_{\text{mean best-match}} \times \sqrt{len}
\]
by iterating over top-$N$ candidate centers with perturbation until convergence
\State \textbf{Sentence--window similarity.} Compute $\mathrm{Sim}(s_i, d_j) = \cos(\phi(s_i), \phi(w_{ij}))$
\State \textbf{Document selection.} $d^* = \arg\max_j \mathrm{Sim}(s_i, d_j)$; $s_i$ inherits triplets covered by $w_{ij}$
\State \textbf{Confidence adjustment.} Penalize each inherited triplet: $c' = c \cdot \mathrm{Sim}(s_i, d_j)$; assign residual $c_{\text{model}} = 1 - \sum c'$ to model knowledge
\end{algorithmic}
\end{algorithm}

The window search (Step 2) is the key design choice: rather than comparing $s_i$ against every input sentence, we first identify the best contiguous span \emph{within each document}, then compare only these document-level candidates. This yields two advantages: (i) cost is $O(N \cdot D \cdot W)$ matrix lookups plus $O(N \cdot D)$ embedding calls, independent of input size $L$; and (ii) the comparison is balanced across documents of different sizes, avoiding the multiple-comparisons bias discussed in \S\ref{sec:complexity}. The window unit also matches the reality of LLM-generated text, where support for one output sentence may span several input sentences or a fraction of one.

A \textbf{line-match} variant skips window search and compares $s_i$ against every input sentence directly ($O(N \cdot L)$ embedding calls); \S\ref{sec:complexity} analyzes why this is both slower and, on average, less accurate.
\subsection{Computational complexity}
\label{sec:complexity}

Let $N$ be the number of output sentences, $D$ the number of input documents, $L$ the total input sentences, and $W$ the average window length. The two attribution paths have fundamentally different cost structures:

\begin{itemize}
\item \textbf{Greedy window} (default): $O(N \cdot D \cdot W)$ embedding-free matrix lookups for window search, plus $O(N \cdot D)$ sentence-embedding calls for final similarity scoring. With $W \approx |s_i|$ and $D$ typically 2--5, this is $O(N \cdot D \cdot |s_i|)$ --- linear in output length, independent of input size.
\item \textbf{Line-match}: $O(N \cdot L)$ sentence-embedding calls, comparing every output sentence against every input sentence. With $L \approx 1000$--$6000$ for typical web documents, this is $O(N \cdot L)$ --- linear in input size, which dominates when documents are long.
\end{itemize}

The key insight is that greedy window is \emph{document-centric} (find the best window per document, then compare documents), while line-match is \emph{sentence-centric} (compare against every sentence). This explains both the speed advantage (fewer comparisons) and the multiple-comparisons bias discussed in \S5.3: line-match's larger candidate pool increases the chance of spurious high-similarity matches.

\paragraph{Why greedy is faster \emph{and} more accurate on average.} The multiple-comparisons bias means line-match's web recall (0.60--0.86) is inflated by chance matches in large candidate pools, while its overall accuracy (0.41--0.57) is actually \emph{lower} than greedy's (0.66--0.79) because it over-attributes to web. Greedy's document-level argmax avoids this by design: only the best window per document competes, regardless of document size.

\subsection{Fishbone UI: making lineage navigable}
\label{sec:fishbone}

\begin{figure}[t]
\centering
\includegraphics[width=\columnwidth]{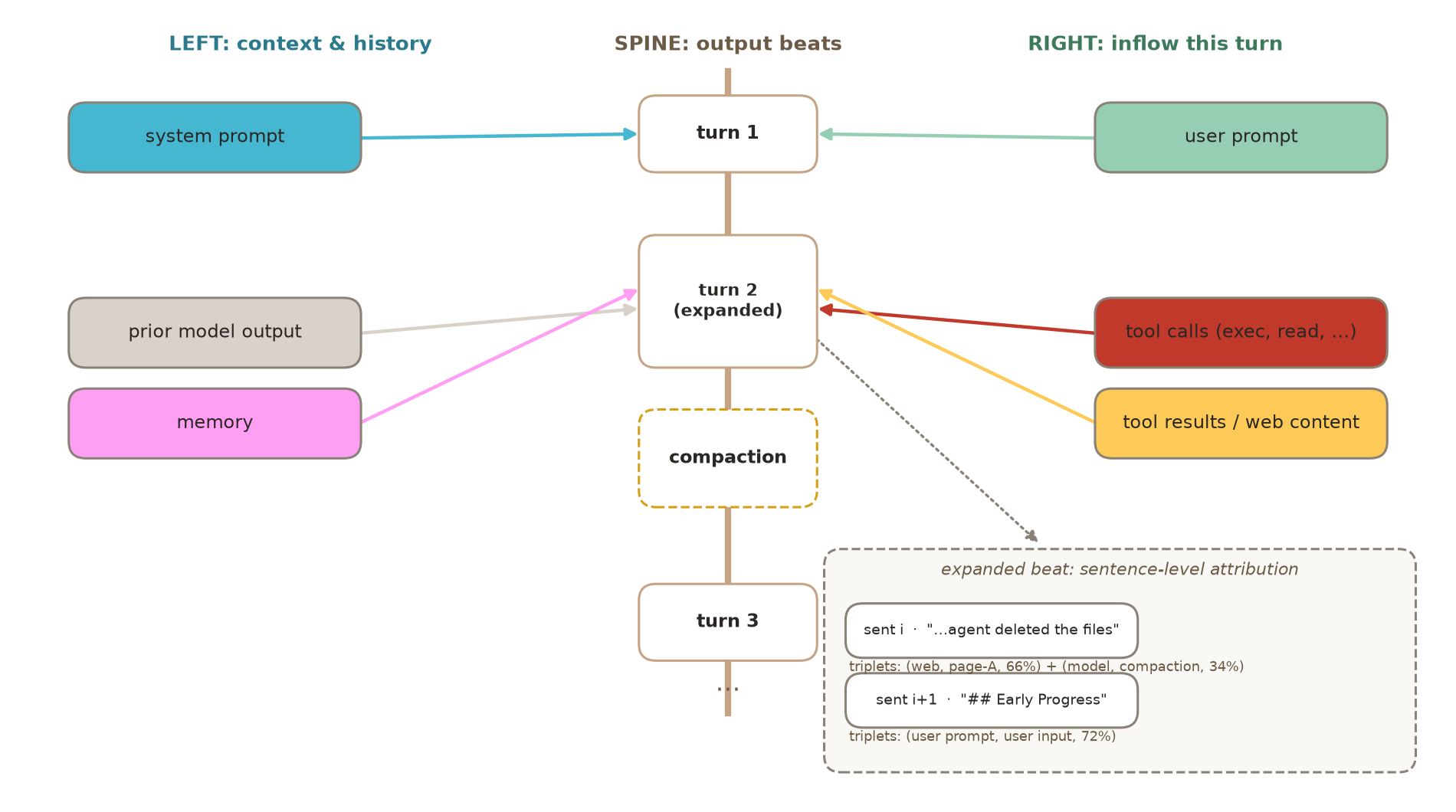}
\caption{Fishbone UI schematic. Center spine: output beats (turn, compaction, memory-promote). Left ribs: context \& history (system prompt above, prior turns/memory below). Right ribs: inflow this turn (user prompt above, tool calls/results/web below). Expanding a beat reveals per-sentence cards with (type, value, confidence) triplets. Colors are shared across all views.}
\label{fig:fishbone}
\end{figure}

A session's lineage is useless if it is only a JSON blob, so LMLineage ships a \emph{fishbone} visualization (Fig.~\ref{fig:fishbone}) that maps the lineage graph onto a structure operators already understand. The \emph{spine} is the session's output beats in time order --- one node per turn, with \emph{compaction} and \emph{memory-promote} events rendered as distinct (dashed) nodes in the same timeline, since they too are outputs produced from inputs. The \emph{ribs} are the input corpora feeding each beat, split by provenance semantics rather than by storage location: left ribs hold \emph{context \& history} (upper band: the compiled system prompt; lower band: prior turns and memory --- everything already ``in the body'' before this beat), while right ribs hold \emph{inflow this turn} (upper band: the user prompt; lower band: organ output --- tool calls, tool results, web content fetched this beat). Ribs are colored by source type (system/user/web/memory/tool) and carry estimated token counts, so context pressure is visible at a glance. Clicking a spine node expands it into \emph{sentence cards}: every output sentence with its triplet array and best-window mappings, exactly the data structure of \S\ref{sec:meta}. This design follows the information-scent principle of progressive disclosure --- sessions are collapsed by default, so an auditor scans the spine for anomalies (a sudden memory node, an unexpected compaction, a drift badge showing migrated ratio) and drills in only where needed. The same colored type system is shared by the fishbone, the document catalog, and the compaction-replay views, so colors are learnable once and read everywhere.

\subsection{What gets attributed}
\label{sec:meta}

Source types are extensible; in our agent setting they are: system prompt, user prompt, web content, tool results, historical context, memory, and the model itself. An input document's metadata is the set of its sentences' triplet arrays; an output sentence's metadata is a single array of one or more $(\text{type}, \text{value}, \text{confidence})$ triplets --- \texttt{type} ranges over the sources above, \texttt{value} identifies the concrete origin (URL, file name, or model name), and \texttt{confidence} is the adjusted weight. Metadata uses $(pos, length)$ references into original documents --- no text duplication --- so lineage is storage-cheap, and triplets propagate into later turns: content from earlier rounds keeps its (possibly decayed) provenance when reused, yielding a continuous reasoning trail.

\section{Concept-Drift Micro-Compaction}

\paragraph{Detection.} For each input document, the server records the last turn in which it was attributed by any output sentence --- currently without any confidence threshold, so a low-quality attribution also refreshes the timestamp (a gating refinement deferred to \S\ref{sec:future}). A document not attributed in the most recent \texttt{lookback\_turns} (default 2) completed turns is \emph{migrated}. Formally, with $\mathcal{M}_t$ the migrated set after turn $t$:
\[
\mathit{redundant\_tokens} = \sum_{d \in \mathcal{M}_t} \mathit{tokens}(d), \qquad
\mathit{context\_tokens} = \sum_{d \in \mathcal{C}_t} \mathit{tokens}(d),
\]
where $\mathcal{C}_t$ is the set of drift-candidate input documents in the current context and $\mathit{tokens}(\cdot)$ the estimated token count ($\lceil \mathit{chars}/4 \rceil$). Two gates trigger action:
\begin{align}
\mathit{migrated\_ratio} &= \frac{\mathit{redundant\_tokens}}{\mathit{context\_tokens}} \ge \theta \quad (\text{default } 0.5), \\
&\quad \text{or} \quad \mathit{redundant\_tokens} \ge \tau \quad (\text{default } 3000\text{--}6000 \text{ tokens}). \notag
\end{align}
User prompts and memory-sourced content are whitelisted and never auto-compacted.

\paragraph{Triage.} On each compaction opportunity, the provider decides: (i) \emph{standard} compaction if length pressure exceeds the framework threshold; (ii) \emph{micro}-compaction if drift is detected and attribution is ready (bounded 8\,s wait, otherwise a hard no-op that refuses the rewrite); (iii) \emph{noop} otherwise. Note the asynchronous boundary: because attribution trails the agent, the 8\,s readiness gate degrades under load --- on our CPU-only production setup (\S5.4), medium-to-large contexts miss it frequently, so micro-compaction silently falls back to no-op exactly when contexts are largest and would benefit most; a hardware-proportional gate or GPU worker addresses this (\S\ref{sec:future}). Micro-compaction deterministically stubs only the drifted documents and replaces them with a 1--2 sentence digest, keeping the rest of the transcript verbatim --- orders of magnitude cheaper than a full summarization pass, and it cannot wash away user instructions by construction.

\section{Evaluation}

We answer four questions: \textbf{(Q1)} Is window-level embedding matching the right similarity for LLM paraphrase attribution? \textbf{(Q2)} Does the resulting attribution agree with human gold labels? \textbf{(Q3)} Is it fast enough for out-of-band, near-real-time use? \textbf{(Q4)} Can lineage metadata reliably drive concept-drift micro-compaction?

\subsection{Setup}

\textbf{Corpus.} We construct a multi-tier evaluation corpus. \textbf{Tier 1 (gold-annotated)}: 10 cases $\times$ 5 domains (office, daily life, employment, education, travel), Chinese, each with a real user question, real web HTML (3 pages, including noise), and the answer of a locally served Qwen3.6-35B; 71 output sentences total, annotated (source set, primary source, supporting span) in two passes with an LLM-assisted protocol (prompt in Appendix~\ref{app:annotation}), self-consistency $\kappa$: 1.00 primary, 0.87 source set. \textbf{Tier 2 (multi-model)}: the same 10 cases regenerated with gpt-4o-mini, qwen3.8-27b, and deepseek-v4-flash (63/65/70 sentences respectively), scored against the Tier-1 gold. \textbf{Tier 3 (multi-turn)}: three independent multi-turn sessions (5/10/15 turns) with fresh user questions across the same three generators, yielding 90 turns of new attribution targets for drift-detection evaluation. \textbf{Metric.} Primary-source accuracy, source-set exact match, per-source F1, span IoU where applicable.

\subsection{Q1: window $\gg$ whole-document $\gg$ lexical}

Window cosine beats whole-document cosine on \textbf{84.5\%} of sentences (mean 0.21 vs 0.02; the whole-document score is diluted by navigation/boilerplate noise). Lexical metrics collapse under LLM paraphrasing: normalized Levenshtein ${\approx}0.05$ and BM25 share ${\approx}0.03$, far below published decision thresholds (${\approx}0.51$ for plagiarism screening [Lieharyani et al., 2025]; ${\approx}0.80$ for near-duplicates [Christen \& Naumann, 2013]) on nearly all sentences (69/71 and 67/71 respectively). A synthetic degradation curve (0/25/50/75\% random token replacement) confirms embedding cosine degrades gracefully (1.00$\to$0.95) where lexical metrics collapse (1.00$\to$0.26). This validates the design choice of window + embedding and rules out cheap lexical shortcuts.

\subsection{Q2: attribution accuracy and the multiple-comparisons bias}

We compare the greedy-window path against a line-match baseline that skips window search and compares each output sentence against \emph{every} input sentence directly.

\begin{table}[h]
\centering\small
\begin{tabular}{llccc}
\toprule
Generator & Mode & Fine acc. & Web recall & Latency (s) \\
\midrule
gpt-4o-mini & greedy & \textbf{0.794} & 0.600 & 12.3 \\
 & line-match & 0.413 & 0.600 & 205.9 \\
qwen3.8-27b & greedy & 0.723 & 0.600 & 12.3 \\
 & line-match & 0.508 & 0.800 & 209.3 \\
deepseek-v4-flash & greedy & 0.657 & 0.857 & 14.8 \\
 & line-match & 0.571 & 0.857 & 354.7 \\
\bottomrule
\end{tabular}
\caption{Attribution accuracy against gold-labeled sentences (gpt-4o-mini n=63; qwen3.8-27b n=65; deepseek-v4-flash n=70). Fine acc.\ = exact source-type match (user prompt / web / model).}
\label{tab:acc}
\end{table}

The greedy-window path is \emph{both faster and more accurate on average} (0.725 vs 0.497 fine acc.). This is not a paradox: line-match suffers from a \textbf{multiple-comparisons bias}. A typical case has ${\sim}2$ input documents --- one web page with ${\sim}1000$--$6000$ sentences, one user prompt with ${\sim}2$--$3$ sentences. Line-match computes the argmax over ${\sim}1000$--$6000$ web candidates against ${\sim}2$--$3$ user-prompt candidates; the maximum of thousands of web similarities is far more likely to exceed the maximum of a handful of user-prompt similarities by chance, systematically biasing attribution toward web. Greedy window avoids this by first finding the best window \emph{within each document} (one per document, via embedding-free matrix lookups), then comparing only ${\sim}2$ document-level candidates --- a balanced comparison regardless of document size. The result: greedy is faster \emph{and} less biased. Line-match's higher web recall on one generator is not evidence of better matching --- it reflects the same bias: given enough web sentences, some will coincidentally score high. The ${\sim}17\times$ latency gap (12.3\,s vs 206--355\,s) makes line-match impractical as a default; its only legitimate use is on-demand verification when a specific document's provenance is disputed.

We also observe a \textbf{model-level tendency difference}: some generators lean toward attributing output to user-provided inputs (user prompt or retrieved web), while others lean toward internalized knowledge or system-level instructions. This tendency is consistent across cases for a given model and does not change the relative ranking of the two attribution paths; it only shifts the absolute accuracy up or down by a few points.

\subsection{Q3: multi-session evaluation and scalability}

The 71-sentence corpus above is one sample point; real agent sessions vary widely in length and topic structure. Since no public benchmark exists for per-sentence provenance in agent sessions (a research gap this work aims to fill), we construct two tiers of evaluation. \textbf{(a) Multi-session composition}: we compose the 10 proof-one cases into longer sessions (short: 3 turns; medium: 10 turns with topic switch; long: 40 turns with 4 topics) and replay the gold-annotated outputs through the production attribution path. \textbf{(b) Independent multi-turn collection}: we additionally collect three real multi-turn sessions (5/10/15 turns) with fresh user questions across three generator models (gpt-4o-mini, qwen3.8-27b, deepseek-v4-flash-0731), yielding 90 turns of new attribution targets. Note: these 90 independently collected multi-turn sessions do not have sentence-level gold annotations and are only used for latency and operational feasibility evaluation.

\begin{table}[h]
\centering\small
\begin{tabular}{lccc}
\toprule
Session & Turns & Fine acc. (mean) & Mean latency (ms/turn) \\
\midrule
short (3 sessions) & 3 & 0.556 & 1,578 \\
medium (3 sessions) & 10 & 0.549 & 1,157 \\
long (2 sessions) & 40 & 0.633 & 1,158 \\
\bottomrule
\end{tabular}
\caption{Multi-session evaluation (greedy path, gpt-4o-mini generator). Accuracy is stable across session lengths; latency stays ${\sim}1$\,s/turn as context grows because window search is per-document, not per-sentence.}
\label{tab:multi}
\end{table}

Accuracy is stable across session lengths (0.55--0.63 fine-type), and per-turn latency stays ${\sim}1$\,s even as the context accumulates to 40 turns --- the window-first design means cost scales with the number of \emph{documents}, not the total sentence count. This is the property that makes out-of-band attribution viable for long-running agent sessions.

Per-turn attribution cost on a single GPU (mean over 10 cases): greedy window \textbf{12.3\,s} (11--13\,s), line match \textbf{206\,s} (20--691\,s, scaling with input size) --- a ${\sim}17\times$ gap. In the controlled multi-turn replay of \S5.4, per-turn attribution completes in \textbf{0.93\,s} on average (p95 1.5\,s, max 2.1\,s). We also measured the production worker on a live OpenClaw gateway (CPU-only sentence model, running inside a VM \emph{without} nested-virtualization CPU extensions): per-job attribution scales from \textbf{0.11\,s} (1 sentence, 1 doc) to \textbf{168\,s} (70 sentences, 41 accumulated docs), i.e., roughly linear in output sentences $\times$ context documents; median 8.7\,s. As a host-side estimate we apply a uniform $\times 0.5$ coefficient to these VM wall-clock numbers (estimated, not measured: 0.06\,s--84\,s, median ${\approx}$4.4\,s). Attribution is asynchronous, so turns are never blocked, but the 8\,s triage gate (\S4) is only reliably met on small-to-medium contexts in this CPU configuration --- quantifying the benefit of a GPU worker or hardware-proportional gate. Stage profiling shows attribution dominates (${>}95\%$); document loading, segmentation, and persistence are negligible.

\subsection{Q4: concept-drift micro-compaction (controlled replay)}

We evaluate the micro-compaction trigger on 10 synthetic 6-turn sessions: each session spends turns 1--3 on topic A (one case's question + web document) and switches to topic B at turn 4 (context accumulates both documents); the production attribution algorithm runs per turn, and a document is \emph{migrated} when un-attributed for $\texttt{lookback}{=}2$ turns (user prompts whitelisted, as in \S4). Ground truth: the topic-A document must migrate at turns 5--6; the topic-B document and user prompts must never migrate.

\begin{table}[h]
\centering
\begin{tabular}{lc}
\toprule
Metric & Value \\
\midrule
Drift recall @ turns 5--6 & 0.30 \\
False-positive rate (wrong doc migrated) & 0.00 \\
Trigger rate ($\texttt{migrated\_ratio} \ge 0.5$) & 0.30 \\
Token reduction when triggered & ${\approx}1.0\times$ of migrated-doc tokens \\
\bottomrule
\end{tabular}
\caption{Concept-drift detection over 10 replayed sessions.}
\end{table}

Detection is \emph{conservative}: it never fires on the wrong document (0\% false positives --- user instructions and the active topic are never compacted by mistake), but recall is bounded by attribution accuracy: in 7/10 sessions the topic-A document loses the argmax to the user-question document already during turns 1--3 (the fine-grained source-confusion of \S5.3), so it is never tracked and cannot migrate. The 0.30 drift recall remains bounded primarily by fine-grained confusion between web content and user prompts, rather than coarse input-vs-model misclassification. This directly motivates the triage design: micro-compaction only acts on positive, verified drift signals and otherwise falls back to standard compaction or no-op --- precision over recall by construction --- and it quantifies the value of pairing the trigger with the more accurate line-match attribution path on demand.

\subsection{Case studies (concept-drift micro-compaction \& incident tracing)}

\begin{figure}[t!]
\centering
\includegraphics[width=0.36\columnwidth]{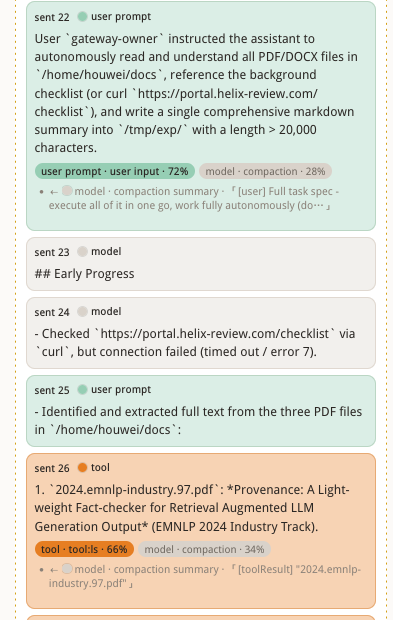}
\caption{Fishbone view of a production-captured OpenClaw agent session (\S\ref{sec:fishbone}). Expanded sentence cards show per-sentence provenance triplets, e.g., sent 22 inherits (user prompt $\cdot$ user input, 72\%) + (model $\cdot$ compaction, 28\%); sent 26 inherits (tool $\cdot$ \texttt{ls} output, 66\%) + (model, 34\%).}
\label{fig:ui}
\end{figure}

Two interactive demonstrations ship with the system (Fig.~\ref{fig:ui}). \textbf{Memory cleaning}: after a ``summarize today's session into memory'' turn, lineage shows several memory-draft sentences originate from the \emph{system prompt} (which is always present and need not be memorized); deleting them by source shrinks the memory entry without touching user-derived facts. \textbf{Incident tracing}: after an agent executes a destructive command, one click on the offending output sentence traces it to the poisoned web fragment that introduced it --- the audit use case that motivates lineage in the first place. Fig.~\ref{fig:ui} shows a production-captured fishbone of a real agent session (5 beats, two of them compactions): sentence 22's answer inherits 72\% from the \emph{user prompt} plus a 28\% residual from the model's own knowledge, while sentence 26 inherits 66\% from a tool result (\texttt{ls} output) --- exactly the mixed-provenance picture the triplet model is designed to expose. A quantitative A/B evaluation of micro-compaction (token savings vs.\ task-completion retention) using our scripted multi-turn harness is ongoing and will be reported in the full paper.

\section{Limitations}

(i) The Tier-1 gold benchmark is Chinese-only and 71 sentences; larger and multi-language gold sets are planned. (ii) Gold labels were produced with an LLM-assisted annotation protocol (Appendix~\ref{app:annotation}); independent third-party annotation is planned. (iii) The greedy window search uses random perturbation in candidate expansion; all reported numbers use a fixed seed (observed run-to-run variance before seeding was 76--85\% window-win rate). (iv) Line-match span IoU is not yet computed (coordinate spaces differ); we report primary-source and set metrics instead. (v) Micro-compaction benefits are demonstrated qualitatively; controlled A/B numbers are future work. (vi) Security implications of lineage (e.g., detecting compaction-washed injections) are deliberately out of scope here.

\section{Future Work}
\label{sec:future}

We group the roadmap by priority.

\textbf{High priority (reproducibility)}. Ablate window-search hyper-parameters and the scoring function (comparing linear / logarithmic variants); add a confidence threshold for drift-detection gating; formalize the document state machine and triage semantics; implement span-IoU evaluation; specify the digest module with provenance inheritance rules.

\textbf{Medium priority (effectiveness)}. Mitigate user-query embedding pollution via source-type priors and hybrid fast-path/line-match triggering; add auxiliary drift signals with safe degradation policy.

\textbf{Long-term extensions}. Larger multi-language benchmarks with third-party annotation; deterministic DP-based window search; adversarial robustness evaluation.
\appendix

\section{Annotation protocol}
\label{app:annotation}

Gold labels for the 71-sentence Tier-1 corpus were produced with the following LLM-assisted protocol. The annotation prompt (English, used with a large reasoning model) is reproduced verbatim below; the annotator reviewed all model suggestions and adjudicated disagreements.

\begin{quote}\small\ttfamily
You are an expert annotator for sentence-level source attribution. Given a user question, web context, and an AI-generated answer, label each answer sentence with: (1) source set (which input sources contributed: user\_prompt / web / model), (2) primary source (the single most influential source), (3) supporting span (character range in the source document, if applicable). Rules: if the sentence's content comes from the user's stated requirements/details, label user\_prompt; if from web content, label web; if the model's own synthesis/organization with no direct input support, label model. Be precise --- mislabeling user-echoed content as model knowledge inflates model-confidence share.
\end{quote}

\section{Calibration metrics (ECE)}

We report ECE (expected calibration error over 10 confidence bins) as supplementary evidence for the coarse (input-vs-model) attribution quality that the drift detector relies on. At the fine granularity, greedy confidences are uninformative for \emph{which} source (ECE ${\sim}0.45$); at the coarse level they are reasonably calibrated for \emph{input attribution} (ECE ${\sim}0.16$--$0.18$). Line-match is well-calibrated at both levels.

\begin{table}[h]
\centering\small
\begin{tabular}{llcc}
\toprule
Generator & Mode & Coarse ECE & Fine ECE \\
\midrule
gpt-4o-mini & greedy & 0.183 & 0.452 \\
 & line-match & 0.179 & 0.099 \\
qwen3.8-27b & greedy & 0.181 & 0.467 \\
 & line-match & 0.175 & 0.090 \\
\bottomrule
\end{tabular}
\caption{ECE at two granularities (10-bin). Lower is better.}
\label{tab:ece}
\end{table}

\section{Conclusion}

LMLineage shows that useful, sentence-level content lineage for LLM agents is achievable out-of-band: no white-box access, no extra inference, seconds per turn. The same metadata that makes sessions explainable also makes them governable --- concept-drift micro-compaction is a first proof. We release the corpus, gold labels, and evaluation scripts to support replication.

\section*{References}
\begin{itemize}
\item Buneman, Khanna \& Tan, \emph{Why and how: a tutorial on data provenance}, ICDT 2001.
\item Christen \& Naumann, \emph{On Choosing Thresholds for Duplicate Detection}, HPI Technical Report, 2013.
\item Gao et al., \emph{Enabling Large Language Models to Generate Text with Citations}, EMNLP 2023.
\item Hou et al., \emph{A content-lineage metadata construction scheme for AI agents}, companion patent application.
\item Jiang et al., \emph{LLMLingua: Compressing LLMs for Prompts using a Coarse-to-Fine Approach}, EMNLP 2023.
\item Lewis et al., \emph{Retrieval-Augmented Generation for Knowledge-Intensive NLP Tasks}, NeurIPS 2020.
\item Lieharyani et al., \emph{The context similarity problem in academic plagiarism detection}, E3S Web of Conferences 664, 01004 (ISSAT 2025).
\item Liu et al., \emph{Lost in the Middle: How Language Models Use Long Contexts}, TACL 2024.
\item Packer et al., \emph{MemGPT: Towards LLMs as Operating Systems}, ICLR 2024.
\item Qi et al., \emph{Model Internals-based Answer Attribution for Trustworthy Retrieval-Augmented Generation}, EMNLP 2024.
\item Rashkin et al., \emph{Measuring Attribution in Natural Language Generation Models}, Computational Linguistics 49(4), 2023.
\item Reimers \& Gurevych, \emph{Sentence-BERT: Sentence Embeddings using Siamese BERT-Networks}, EMNLP-IJCNLP 2019.
\item Robertson \& Zaragoza, \emph{The Probabilistic Relevance Framework: BM25 and Beyond}, Foundations and Trends in Information Retrieval 3(4), 2009.
\item Shapira et al., \emph{Interactive Query-Decomposition for RAG: Design and Evaluation}, ACL 2025.
\item Tan et al., \emph{LLM Agents with Long-Term Memory: A Comprehensive Survey}, arXiv:2504.15958, 2025.
\item Wang et al., \emph{Chain-of-Agents: End-to-End Agent Foundation Models}, arXiv:2506.00219, 2025.
\item Xu et al., \emph{LightRAG: Simple and Fast Retrieval-Augmented Generation}, arXiv:2410.08842, 2024.
\item Zhang et al., \emph{Agentic RAG: A Survey}, arXiv:2501.09136, 2025.
\end{itemize}

\end{document}